# Stable Parallel Looped Systems – A New Theoretical Framework for the Evolution of Order[1]

Muralidhar Ravuri[2]

## Abstract

The objective of the paper is to identify laws and mechanisms that allow the creation of more order from disorder using natural means i.e., without the help of conscious beings. While this is not possible for the collection of *all* dynamical systems as it violates the second law of thermodynamics, I show that this is possible within a special subset called stable parallel looped (SPL) dynamical systems. I identify a new infinite family of physical and chemical dynamical SPL systems, which are (a) easy to create naturally and (b) easy to merge, link and combine to create dynamical systems of any specified complexity. Within SPL systems, I propose a special collection of designs called active material-energy looped systems using which it is possible to generate large-scale ordered chemical networks, like the metabolic networks, in a reliable, repeatable, iterative and natural manner. The resulting SPL systems provide a new theoretical framework for the problem of origin of life.

---

# Introduction

Consider a sealed room filled with several types of simple molecules like $CH_4$, $NH_3$, $H_2O$, $H_2O_2$, $H_2S$, $CO_2$ and others. Let us assume that it is supplied with sufficient amount of randomly distributed energy for as long as we are performing this experiment. Initially, the molecules in the room are in a state of complete disorder. The question we ask now is: what would we expect to see in this room after, say, 100 million years? If the same set of molecules stays in the same state of disorder, we clearly do not require any further explanation. On the other hand, if we see considerable order like L-form amino acids, D-form sugars, DNA, enzymes and even primitive life forms, we have a difficult question to answer. How did such high degree of order get created from the initial disorder? What natural laws and mechanisms make this happen, assuming there is no help from conscious beings? The only answer for now is random chance-based mechanisms, which is clearly unsatisfactory for the high degree of order. The purpose of this paper is to present a rigorous alternative by identifying new structures, dynamics and less chance-based mechanisms to create order from disorder.

Take, for example, the familiar notion of stability of molecules. Those molecules that are stable under a given set of environmental conditions (like temperatures and pressures) remain after a long time while the unstable molecules disintegrate and disappear (or become trace elements like $H_2O_2$ compared to $H_2O$). Therefore, in the above sealed room, we see that stability alone is capable of producing order from disorder. After 100 million years, only those molecules from the initial set that are stable will remain. We want to know if there are additional patterns and properties beyond stability that can bring further order within this sealed room.

If we place highly ordered systems like enzymes and bacteria into the sealed room as the initial set, it is not surprising that more order will be created subsequently (see Darwin's theory). Therefore, the above problem of evolution of order is interesting only if the initial structures in the sealed room are the most elementary ones. I refer to them as *minimal structures*, using which we want to explain how nature creates highly ordered systems.

Consider another example of order, namely, the long ordered sequence of chemical reactions that occur within our body at every instant for every task (like when walking or when digesting food). If we place the same collection of molecules in a beaker, as in a dead cell, the same cascading sequence of reactions cease to occur. What static and dynamical arrangement of molecules allows such coordination to occur in living cells but not in dead cells? The basic survival of an organism requires such coordination. While we have abundant proof for the physical existence of such high degree of order (i.e., every living organism), what we do not have are the theoretical structures and mechanisms that explain the creation of large-scale stable complex systems like living beings.

From these examples, we have two distinct questions to answer. One is an *analysis* problem in which we want to understand the laws and mechanisms that allow the existence and evolution of large-scale highly-ordered complex systems. The second is a *design* problem in which we want to create such large-scale complex systems both naturally and synthetically. Every existing theory that attempted to provide a framework for such problems started analyzing and designing a system with a small degree of order. However, they failed to scale to systems involving thousands of reactions or dynamical subcomponents. As a result, there is a huge gap in the frameworks used by theorists and experimentalists in biology and complex systems. Experimentalists argue that the models chosen by theorists are not realistic and they do not capture the scale of complexity observed within living cells.

The central feature of this paper is the presentation of a new theoretical framework that allows the creation of large-scale (millions or, in fact, an infinite number) and highly interconnected dynamical systems with 'any' specified complexity, which as a whole does not collapse for a sufficiently long time. This new class of dynamical systems called stable parallel looped (SPL) systems addresses the above discussed issues and acts as a common framework for both theorists and experimentalists. They not only capture the complexity of observed reality of living systems, but they also let us create complex *dynamical* systems that cannot form naturally.

Let me formally state the objective of this paper as follows: *describe the structures, patterns and natural mechanisms within existing physical laws using which the probability of creating large-scale stable dynamical systems with high-order from less ordered components keeps increasing over time*. A few points to emphasize are: (a) in addition to synthetic mechanisms, with help from conscious beings, we should provide natural mechanisms for the evolution of order, (b) the theory should allow analysis of existing large-scale systems like cells as well as design of new large-scale synthetic dynamical systems and (c) the probability of creation of order should increase over time – this eliminates pure chance-based mechanisms. For example, there is no increase in probability of bringing the molecules in a small enough region within a short interval, a necessary requirement before we can create order, in the second million years after, say, 'randomly' trying for the first million years. It is apparent that the problem of origin of life is a subset of the above problem of evolution of order. The less ordered components are inanimate objects and the high-order stable dynamical systems are the primitive life forms.

For the above problem of evolution of order, one fundamental question typically ignored or postponed is the following. When a highly-ordered system is created naturally, we focus on identifying the minimal structures and sequence of steps that caused the creation of order. However, what laws and mechanisms *guide* all necessary subcomponents to indeed follow these identified sequences of steps? For example, according to RNA world hypothesis the minimal structure for the origin of life is the creation of an RNA replicase enclosed in a closed membrane (*1*). However, to create replicase molecules, several prerequisite molecules and steps are needed. These are, say, lipids, nucleic acids, sugars, arrangement of monomer chains to create unique polymer sequences, inorganic catalysts like clay montmorillonite and others (*2, 3*).

The ignored question translates to (see Fig. S1): (i) how can all of these prerequisite molecules naturally accumulate in a small spatial region (cf. air molecules reaching one corner of the room) so they get a chance to react chemically? This seems to oppose the second law of thermodynamics. (ii) How can they accumulate within a short period? Otherwise, they may disintegrate or move away before they can be part of a long cascading chain of chemical reactions? For a few reactions, abundance of molecules or chance mechanisms as an answer may seem satisfactory (like the spontaneous self-assembly of phospholipids to form a closed membrane). However, this is unsatisfactory for processes of life as they involve long cascading chain of reactions like with creating a unique RNA replicase sequence. (iii) If structures like RNA replicase are created by chance, how can they be created reliably and repeatedly so the order can be recreated in the event they are starved (like during ice ages or meteor strikes) or they disintegrate? The mere presence of all these prerequisite molecules is not enough to create high degree of order (cf. dead cells or all required chemicals dumped in a beaker). There seems to be a natural sense of regulation or coordination at a large-scale that avoids collapse, though it appears highly likely, in living beings. The above questions are applicable even for processes that occur in our body at this instant.

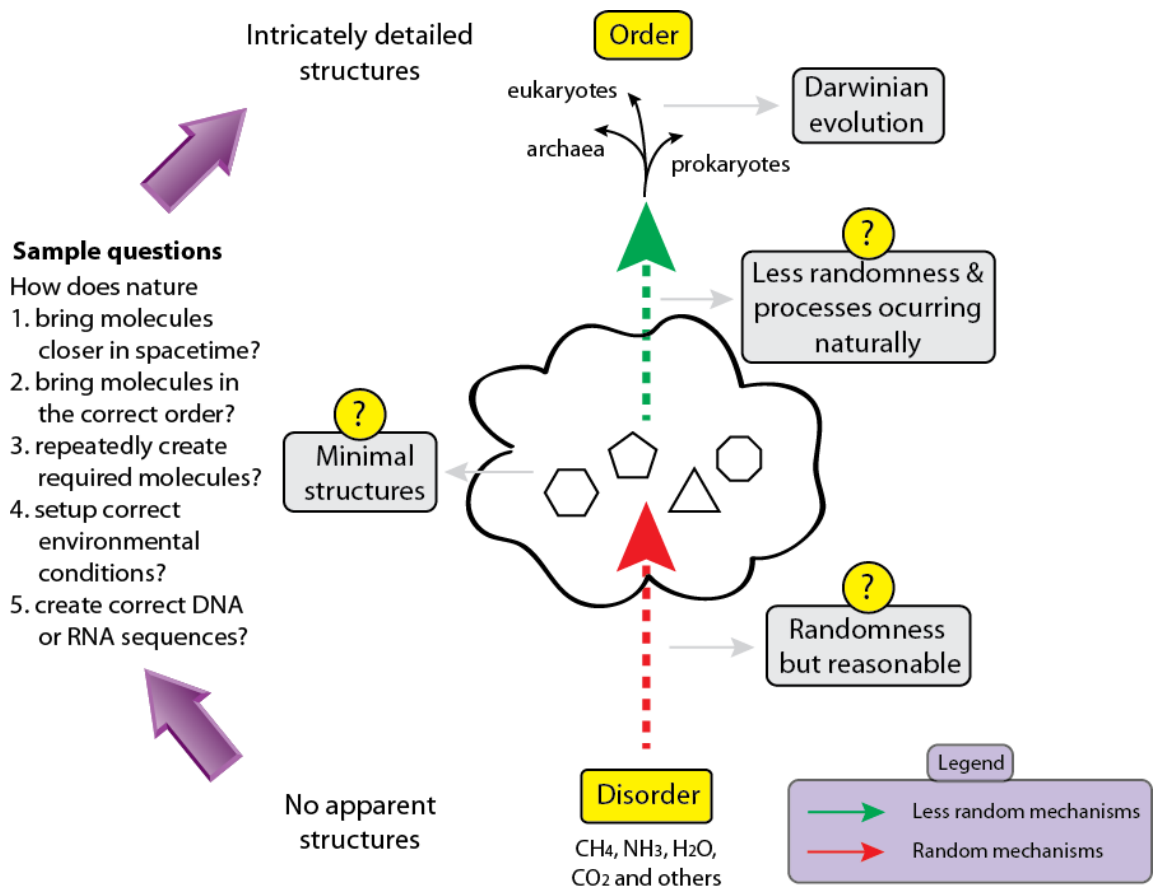

**Figure S1 | The problem of evolution of order.** Identify the minimal structures used to create a threshold system capable of Darwinian evolution with increasing probability. The items marked with '?' need to be identified or clarified. The minimal structures themselves rely on randomness to a degree. However, subsequently, we expect to rely less on randomness. Besides, since intricately detailed structures are created from no apparent structures, we need to answer several questions listed here that are unique to the generation of order.

## Minimal structures

Rather than starting with highly ordered structures to study the evolution of order, we want to start with the simplest of them, namely, the minimal structures. In this paper, I identify four of them using which I show how to create an infinite family of dynamical systems with any specified complexity.

   *Stability*: The first already well-known minimal property is the stability of molecules (with long half-life). A system steadily accumulates nonrandom stable collection of molecules even if we did start with a random collection of stable and unstable molecules. Some of these stable molecules are $CH_4$, $NH_3$, $H_2O$, $H_2$, HCN and others used in prebiotic experiments like Miller-Urey (*4, 5*), Oró (*6*) and others (*7-9*).

***Closed compartment***: The second well-known structure is a closed compartment. Reactions occurring within a closed compartment have better conditions like temperatures and pressures than the ones occurring in the open environment. To create this structure naturally, it was shown that the synthesis of lipids and fatty acids, important for forming closed compartments, can occur under prebiotic conditions (*7-9*). Using these molecules, it was shown that vesicles do form under prebiotic conditions using fatty acids, organized as micelles (*2, 3*). Clay montmorillonite catalyzes the vesicle formation and the vesicles are stable for days to months. In addition, it was shown that when fatty acid micelles are added, the vesicles even grow and split using a process called flipping (*2, 3*).

In spite of an improvement when using a closed compartment over an open environment, the products are generated only if all reactants are able to enter this small spatial compartment within a short period in the correct cascading linear sequence. This is unlikely analogous to air molecules reaching one corner of a room. Therefore, we need additional structure to offer better than random chance-based passive and diffusion mechanisms.

***Active dynamics***: The third minimal structure is an active power source, not just the equivalent of ATP, which powers internal reactions, but more importantly to suck inputs into the closed compartment from the environment. All systems relying on a collection of specific inputs require this feature, at minimum. Indeed, we see that all living beings did retain this feature (as they actively fetch the necessary inputs or food). Fig. S8 described later is a minimal 'pump' design that achieves this goal. With several active suction-based pumps, chemical reactions no longer need to wait for all reactants to passively accumulate in a small spatial region within a short interval by chance or wait for better environmental conditions to appear randomly. The pump (Fig. S8) is the first new structure identified in this paper for the purpose of creating order causing the new theory to deviate from all existing theories on, say, origin of life.

Next, even if favorable chemicals (like $O_2$, ATP and others) are created naturally using the three structures identified so far, how can we address the steady accumulation and repeatable creation of these molecules? The current models account for this by trying to identify unique conditions when the half-life is high. For example, at 100 °C, half-life of decarboxylation of alanine is 19,000 years (*10*). However, molecules like nucleosides, peptide bonds and phosphate esters only have a half-life ranging from a few seconds to a few hundred minutes (*10*). Therefore, expecting the same conditions to work for most molecules is unlikely. Yet, creation of high order requires not only the stable molecules, but the less stable and low half-life ones as well.

***Looped dynamics***: The fourth structure needed is a way to continuously regenerate molecules, namely, a looped set of reactions (analogous to metabolic networks (*11*)). Each reaction within a chemical loop can be assisted by enzymes or using a collection of active pumps mentioned above (Fig. S8). When the loop is completed, all chemicals involved self-sustain and have steady concentrations (i.e., are self-regulated). The looped system as a whole requires abundant energy and chemicals as inputs. In addition to chemical loops, physical loops like circulation, water cycle, seasonal cycles and vortices cause steady mixing to generate diversity. These looped dynamical structures are the second new set of structures identified here unlike all existing theories on the evolution of order. Indeed, we see that living beings have retained the looped dynamical property through the creation and growth of complex metabolic networks (*11, 12*). In the next section, I show how to create an infinite family of physical and chemical dynamical systems with any specified degree of complexity using loops powered by pumps. In fact, the problem of evolution of order now translates to the problem of creating increasingly complex looped networks.

Note that concepts similar to 'chemical' looped systems like hypercycles were studied previously and applied to the problem of origin of life, though they were specialized to autocatalytic systems with self-replicative catalysts (*13, 14*). However, the creation of an infinite family, unification of both physical and chemical SPL systems and design of SPL systems using minimal structures like pumps instead of self-replicative units like enzymes and replicases, all of which will be discussed here, is new.

To see that loops are necessary for the creation of order, consider Fig. S2 which show the energy change for a few sample reactions. Some reactions proceed naturally from an uphill energy state to a downhill state. If after reaching the downhill state, the low-energy molecules do not react and turn into higher-energy molecules by consuming energy, the system, as a whole, would eventually settle down at the lowest energy states (near absolute zero temperature). The only way to avoid this is by supplying external energy, thereby completing loops (Fig. S2). There are multiple ways to bring the molecules uphill – passive random-chance mechanisms, enzymes, replicases or other catalysts (which are non-minimal structures primarily used by all existing theories) and active looped mechanisms with pumps (i.e., the minimal structures used by the new theory).

Using the four structures identified here, the primitive world would start to exhibit detailed chemical networked structure over time. However, as the looped network begins to create polymers, a new set of problems appear. For example, creating random proteins or RNA sequences from a pool of amino acids and nucleic acids have little purpose unless the polymer sequences are precise. What laws or structures direct the creation of a specific polymer sequence for the first time and repeatedly? With RNA world hypothesis, these are chance mechanisms to create an RNA replicase (*15, 16*). If the mechanisms are based on 'parallel' random collisions, creating a randomly permuted RNA sequence and a correct sequence are indistinguishable even though the functional differences are significant. The 'usefulness' of a polymer sequence does not make it any easier to 'create' it reliably and repeatedly. Therefore, we need a 'serial' mechanism using a template sequence. How can nature do this without the equivalent of an RNA replicase, enzymes and other non-minimal structures? In a subsequent paper, I will suggest how additional structures (**helical geometry and complementary nucleotide bases**) can be combined with active dynamical pumps mentioned above to provide a theoretical design/mechanism to solve this problem.

Therefore, the central result of the paper is that using only the minimal structures identified in this section (collectively termed as active stable parallel looped systems), I show (a) how to create order of any specified complexity and (b) how to understand existing ordered systems among both living and nonliving systems.

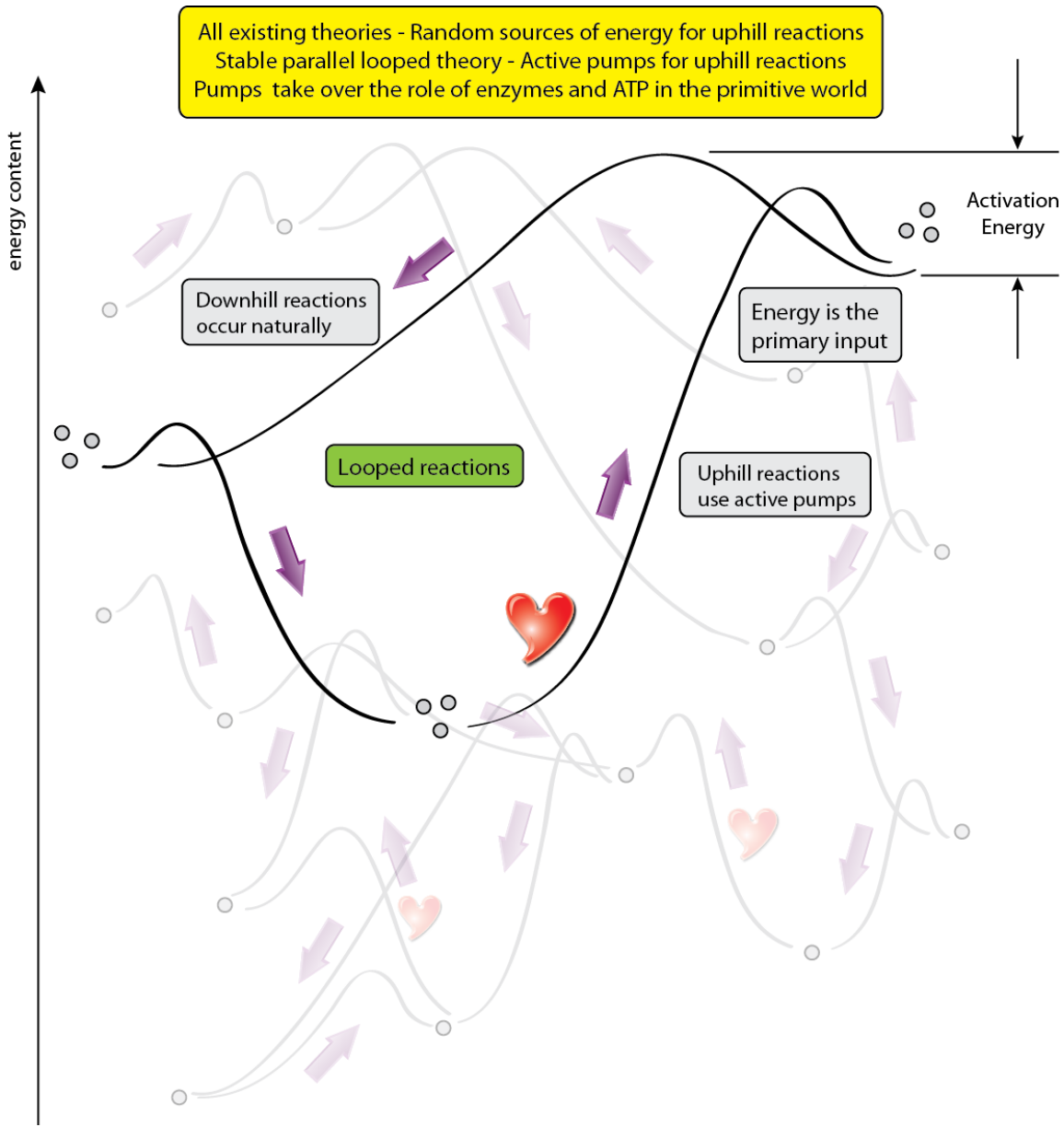

**Figure S2 | Stable parallel looped network with uphill and downhill reactions.** If we represent different molecules at different heights based on their energy content, chemical reactions proceeding downhill can occur naturally. However, reactions proceeding uphill require either passive random energy or active energy. All existing theories on evolution of order assume random energy source for uphill reactions while the new theory uses simple special pumps instead. With active pumps, self-sustainability, reliability, repeatability and continuity can be guaranteed. These simpler, easy-to-create and self-replicating SPL pumps are analogous to enzymes and ATP.

## Stable parallel looped systems

Among the minimal structures above, active pumps and looped dynamical systems are the new ones introduced in this paper. In this section, I will focus on looped systems. I will show how to create order of any specified complexity using an infinite family of looped dynamical systems.

***Definition 1***: A dynamical system is said to be a stable parallel looped (SPL) system if and only if the following conditions hold: (a) *existence of dynamical loops* – there exists parts of the physical or chemical dynamics of the system that recur approximately the same with a time-varying period *T(t)* for a sufficiently long time and, if necessary, when supplied with energy and/or material inputs repeatedly, (b) *parallel interacting loops* – the system has one or more loops that exchange inputs and energy with one other and (c) *stability* – there exists sufficiently small disturbances for which the system does not collapse (like, say, bounded inputs producing bounded outputs).

In order to exclude systems that die down quickly, I will typically impose the following condition as well – an SPL system should be stable sufficiently longer than the time period of the slowest looped dynamics (say, at least 10 times longer). Furthermore, I only require approximate repeatability and periodicity with no necessity to continue the dynamics indefinitely.

The simplest example of an SPL system is a swing in a playground. A person pushes at regular intervals to supply small energy to continue the oscillatory (or looped) motion. The other two conditions in the definition are trivially satisfied. Other examples are limit cycles, periodic, cyclic and oscillatory dynamics (like pendulums and waves). With chemical systems, examples of SPL systems are chemical oscillatory reactions like Belousov-Zhabotinksy reaction, Briggs-Rauscher reaction and interconnected cycles like Kreb's cycle, Calvin-Benson cycle and more general metabolic networks (*11, 12*) within living beings.

Our objective now is to show how to create large-scale physical and chemical SPL systems starting from the simplest SPL systems. I will first present an infinite family of 'synthetic' SPL designs of any specified complexity. They will be linearly ordered from lower to higher amounts of self-sustaining times. Next, I will show how to redesign this family so they can form 'naturally'. Together, they will be seen as a way to create more order from disorder.

***Physical SPL systems***: Consider balls rolling down an inclined plane, which is not an SPL system. One approach to convert it into an SPL system is to include a conscious human who will bring the balls back on top of the inclined plane (Fig. S3(a)). To eliminate the involvement of a human and thereby improving the creational difficulty partially, Fig. S3(b) shows a modified SPL design that curls the inclined plane to form a structural loop. If the energy lost through friction is small, random energy bursts can occasionally cause the balls to reach the top and continue the dynamics for a few loops. The creation of the structural loop of Fig. S3(b), though less likely to occur naturally, is only needed once unlike Fig. S3(a) with a human involvement.

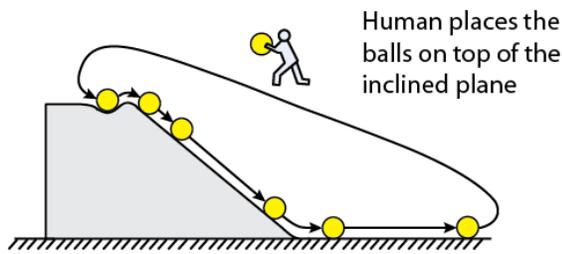
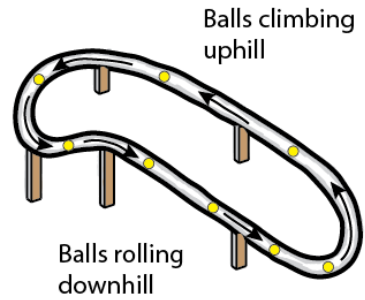
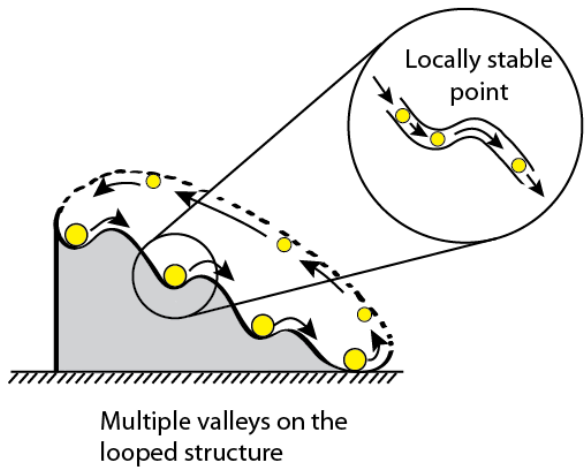
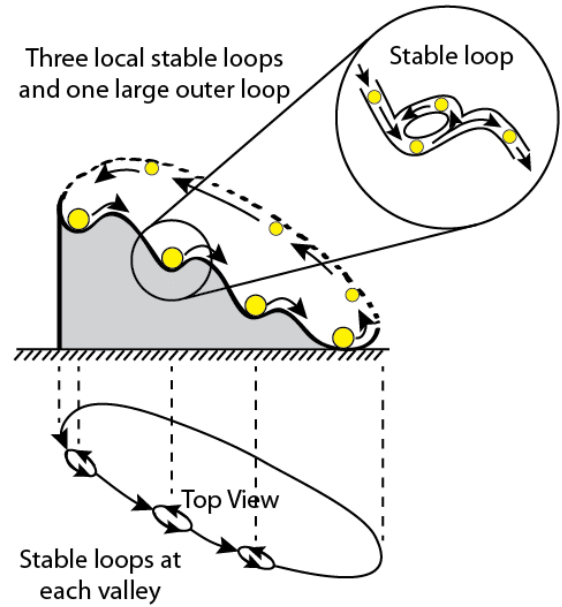

**Figure S3 | A simple set of physical SPL systems.** We consider a simple SPL system in which balls roll down an inclined plane while a human brings it back to complete the loop. We modify this system incrementally by eliminating the human and by adding more complexity. There are several ways to modify the system. I show one in which more locally stable points are added, which are then converted into locally stable loops.

One problem with Fig. S3(b) is that the balls move in both forward and reverse direction. A single stable point at the bottom is the cause of this attraction from both directions. The resulting collisions hinder the overall motion with all balls settling at the bottom (the lowest-energy state). To restart the looped dynamics once again, every single ball needs a sufficiently large amount of energy. Fig. S3(c) is a redesign with several hills and valleys along the looped path (cf. roller coaster designs). With multiple stable points (valleys), the balls store part of the external energy as potential energy at these intermediate heights unlike Fig. S3(b). Restarting the SPL dynamics becomes easier as all balls do not settle at the bottommost valley. The balls also have less average number of collisions as the collisions become local to each stable point. Therefore, this SPL system exhibits dynamics longer than the previous design since the occasional random energy from the environment is both effectively trapped and utilized to produce dynamics.

To decrease the losses further, Fig. S3(d) shows another SPL design in which the stable points are replaced with tiny and well-lubricated stable loops. The balls entering the loop continue their looped dynamics in a single direction avoiding back-and-forth collisions. In addition, looped dynamics avoids the equivalent of high activation energy necessary to restart the motion, had each ball halted its motion. This lowers the dependency on the external world. Small random bursts of external energy, which always exists, say, from wind and other disturbances, would not have been utilized effectively by the previous designs unlike the new design as it has continued dynamics. The extra energy needed to kick the balls to the next higher level is not just from the local bottommost location as in Fig. S3(c), but from any other intermediate location as well, due to their continued motion within the looped path. If there is competition or a genuine shortage of energy, designs similar to Fig. S3(d) (like atomic orbital models with similar transitions between levels and molecular orbital dynamics) will exist much longer than the other designs.

From the above description, we can not only create a large family of physical SPL systems but we can also iteratively improve the operational inefficiencies of these designs. The generic process (Fig. S4(a)) is as follows: (i) introduce a stable loop which typically has one physical stable point, (ii) change the single physical stable point into multiple stable points, (iii) change one or more stable points into stable loops and (iv) continue step (i) for each stable loop iteratively. Fig. S4(a) shows the evolution of one SPL design after two iterations.

It is now possible to merge, link or chain (Fig. S4(b)) multiple SPL systems in an infinite number of ways to create new SPL systems with an increased set of loops and an increased ability to exhibit interesting behaviors. During this process, we need to ensure that the parallel loops are coordinated to have a single direction of flow for achieving stability for a long time. Blood circulation systems and our brain networks are examples of similar physical SPL systems with directed flows. In this way, we can iteratively create stable physical dynamical systems of increasing order and any specified complexity.

Most classes of interconnected dynamical systems have the property that the higher the complexity, the easier it is to destabilize them. SPL systems are the only known class for which the opposite can be satisfied – higher complexity makes them easier to self-sustain longer (cf. Figs. S3(a) - S4). However, the above physical SPL systems cannot be created naturally. To circumvent this difficulty, we need to switch to chemical SPL systems.

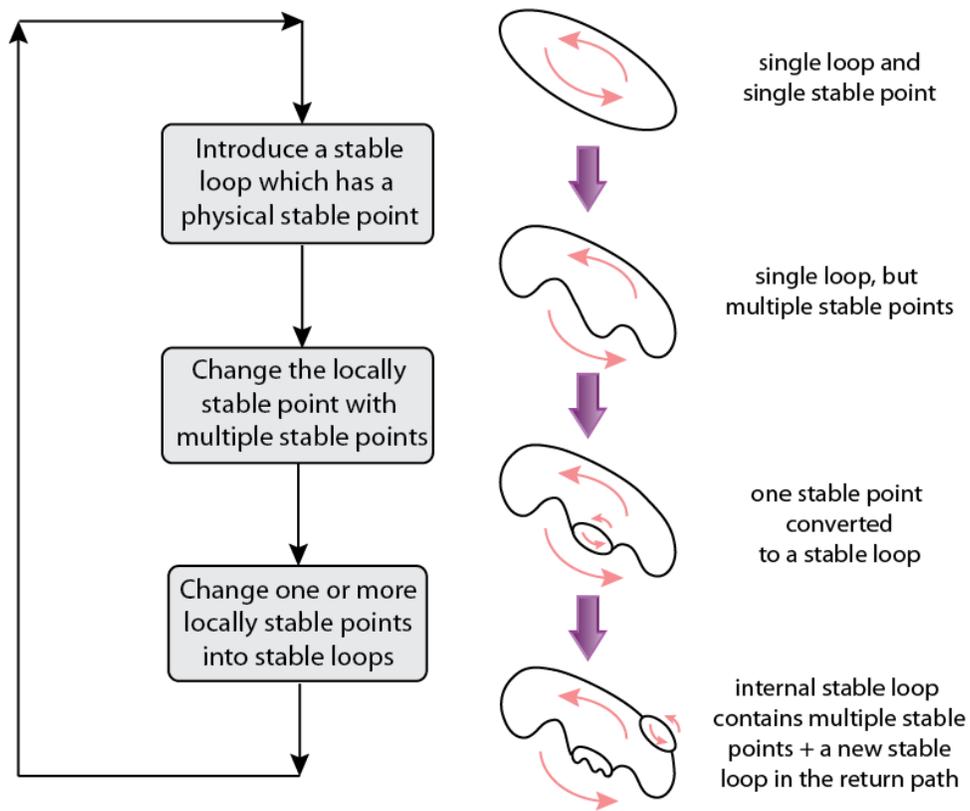
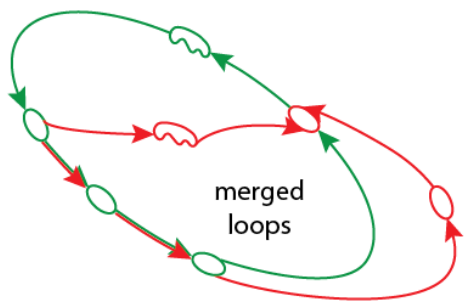
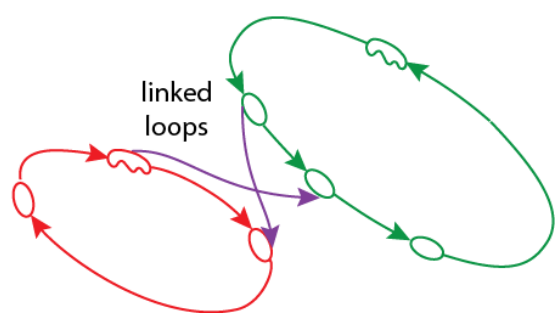

**Figure S4 | An iterative process to create an infinite family of physical SPL systems with increasing complexity.** Starting from a single looped SPL system, it is possible to create a multitude of physical SPL systems depending on how we modify subcomponents of it at various stages of evolution. We can either merge or link two or more physical SPL systems to create a new SPL system with different features and properties.

***Chemical SPL systems***: Given the above infinite family of physical SPL systems, we can show the 'existence' of a large family of chemical SPL systems via a mapping specified in Fig. S5. Here, we map all the balls (considered the same for physical SPL systems) as 'different' types of molecules. Represent different heights of the inclined plane as different energy levels. A ball at a given height is represented by a specific molecule with a corresponding energy state. The act of balls rolling downhill represents a chemical reaction in which reactant molecules of higher-energy state transition to product molecules of lower-energy state (Fig. S5). Though the balls in physical SPL systems do not change, the corresponding representation as molecules in a chemical SPL system do change with each transition. The valleys correspond to special locally stable molecules of a given energy level. The hills represent the activation energy that a chemical reaction needs to overcome.

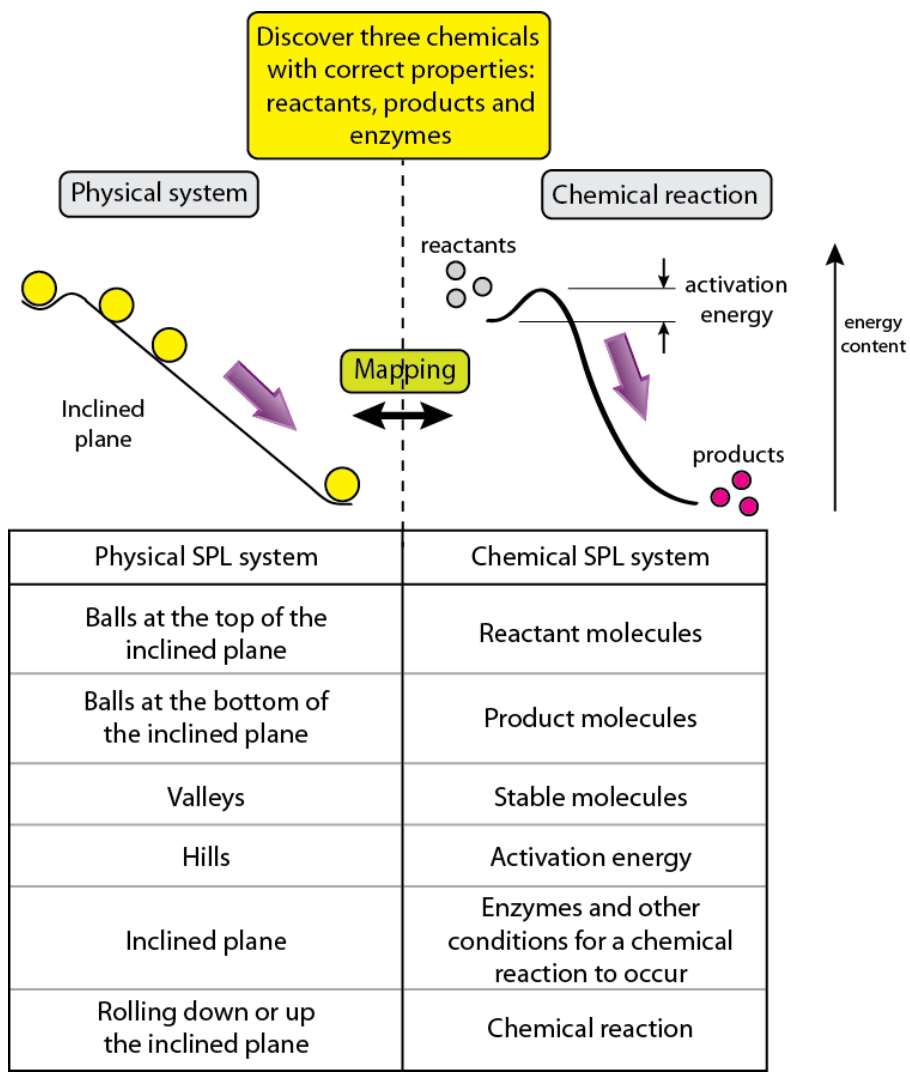

**Figure S5 | Mapping a physical SPL system to a chemical SPL system.** Balls rolling down an inclined plane from one energy state to another are mapped to an uphill or a downhill chemical reaction. As shown in the table, the activation energy, the enzymes and other conditions that assist a chemical reaction are mapped, provided reactants, products and enzymes with suitable properties exist or can be discovered.

There are multiple ways to map the static physical structure of inclined planes in the chemical world. In one representation, they correspond to dynamical enzymes (Fig. S5). Enzymes assist a given uphill and downhill chemical transition. While the balls absorb the external energy directly to produce the motion along the inclined plane, with chemical SPL systems, the enzymes (analogous to the inclined plane itself) also absorb the energy. Their unique shape and size cause the uphill or downhill transition to create new products (cf. induced-fit model of enzymes). In another representation, they correspond to suitable conditions like temperatures, pressures and energies that allow a given chemical transition. Other representations include closed compartmental structures, catalysts, pumps (Figs. S8) and a combination of these. Each of these representations operates in a similar way, though the degree to which these structures assist the uphill and downhill reactions differ compared to enzymes.

Fig. S6 shows one example of chemical SPL system utilizing the above mapping. It contains two downhill reactions and one uphill reaction to complete a chemical loop. The chemical loop itself is virtual unlike a physical loop. The downhill reaction tends to occur easily and naturally while the uphill reaction requires considerable external energy and several favorable conditions, all of which are nontrivial to setup naturally, reliably and repeatedly. In Fig. S6, enzymes assist in converting molecules both in the downhill and uphill directions to form a simple chemical loop. An example of a complex organic chemical SPL system involving multiple loops is Kreb's cycle.

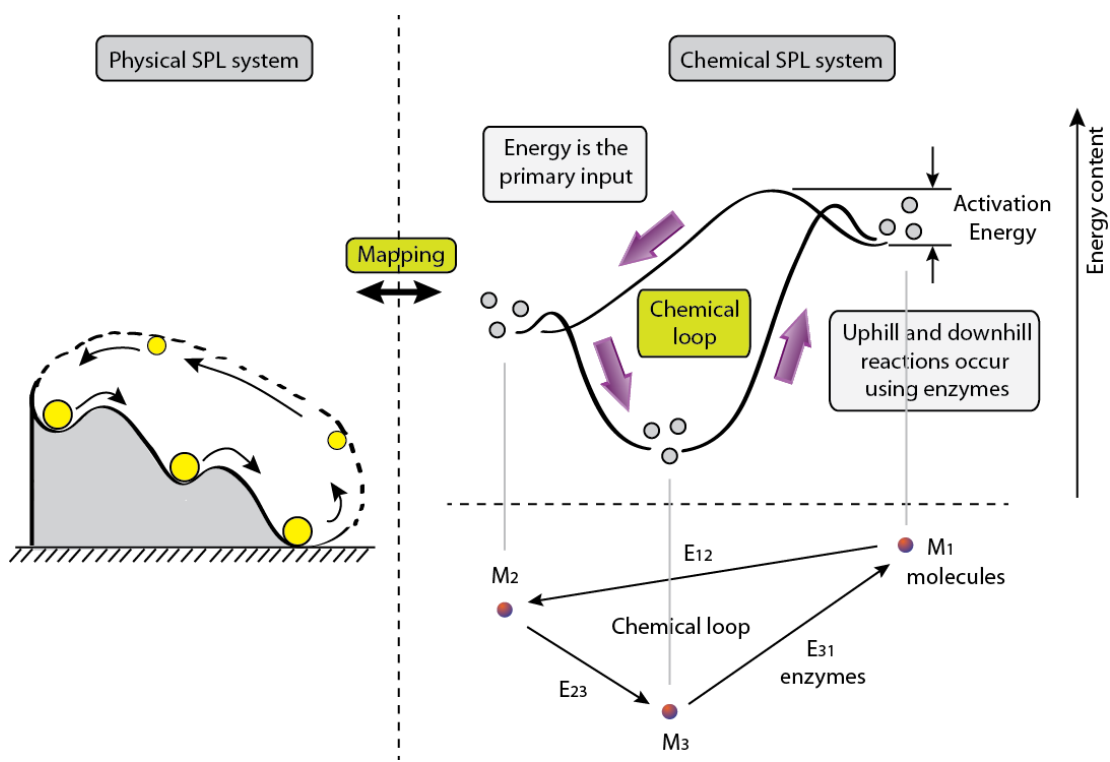

**Figure S6 | A sample chemical SPL system mapped from a physical SPL system.** Using the mapping of Fig. S5, a physical SPL system with three valleys is mapped into a corresponding chemical SPL system with three types of molecules at three energy states. Most metabolic pathways within living beings (like the urea cycle and the Kreb's cycle) are much more complex than this sample chemical SPL system. Each reaction requires the discovery of a suitable enzyme, say, both in the uphill and downhill directions to complete a loop.

From Figs. S5 and S6, mapping a physical SPL system into a chemical SPL system is quite simple in an 'abstract' sense. However, creating a 'real' chemical SPL system is nontrivial because we need to discover different types of chemicals and enzymes with several relative constraints between them. As an example, consider Fig. S6. We need to represent three reactions for the mapping of the physical SPL system. From Fig. S6, each reaction requires three chemicals – reactants, products and enzymes – a total of nine types of chemicals. However, since the physical and chemical reactions are interconnected, they impose constraints – the products of one reaction = reactants of another reaction. This implies we only need three chemicals ($M_1$, $M_2$ and $M_3$) and three enzymes ($E_{12}$, $E_{23}$ and $E_{31}$). Furthermore, the relative energy levels should also be such that $M_1 > M_2 > M_3$. Besides, the activation energies of each reaction (which are altered by enzymes) should be comparable to the corresponding depth of valleys in the physical SPL system that is being mapped.

For more complex physical SPL systems (like even Fig. S4), the practical difficulty of discovering a real chemical SPL system that satisfies all constraints of the correct mapping gets compounded. Having a repository of metabolic networks (*11*, *12*) from each species greatly helps us in finding chemicals with required relative relationships. Each design in Figs. S3-S4 can now be represented as chemical SPL systems using the above mapping with a suitable choice of chemicals and enzymes, if they exist.

Merging, linking, chaining and the generic process mentioned above for physical SPL systems continue to be applicable here (Fig. S7). They help generate a large collection of chemical SPL systems with an appropriate choice of enzymes, organic reactants and products. This collection can be ordered with increasing complexity (using, say, the degree of the chemical network as the measure). As an example, the equivalent of replacing stable points (Fig. S3(c)) with stable loops (Fig. S3(d)) is glucose and $O_2$ forming $H_2O$ and $CO_2$ being replaced with a stable looped reaction, namely Kreb's cycle (Fig. S7(a)). In general, if we pick all the chemicals from the metabolic network (*11, 12*) of a given organism, we get a chemical SPL system. Therefore, *each living being can be viewed as a chemical SPL system*. Even though the above mapping is direct with organic reactions, we can, nevertheless, represent inorganic chemical loops like Belousov-Zhabotinsky and Briggs-Rauscher reactions as interacting SPL systems using the above mapping.

***Self-replicating property of SPL Systems:*** Chemical SPL systems exhibit several properties that help us understand how order comes from disorder. The first of these are a self-replication property, which is a feature unique to life. To see this, consider a collection of molecules and let us ask which subset remains after a long time. The typical answers are (a) inert, less reactive chemicals and ones with high half-life, (b) abundantly available chemicals and (c) chemicals that switch repeatedly and randomly between states. However, with the discovery of chemical SPL systems, we now have a new 'dynamical' case as well.

***Claim 1:*** Any collection of molecules that form a chemical SPL network while consuming (i) freely available external inputs/energy or (ii) molecules that are themselves part of other external loops, will remain with high probability after a long time.

Molecules within the looped subset (a) reuses raw materials while consuming only abundant energy or chemicals unlike nonlooped subset, which produce non-reusable products and (b) self-sustain longer compared to nonlooped dynamics, which die down quickly. Additionally, nonlooped chemicals arriving within a small neighborhood at the correct time as coincidences so they could react, has near-zero probability (cf. air molecules reaching one corner of a room). Looped chemicals, on the other hand, are repetitively produced. As their concentrations keep increasing in a given region, the chances of coincidences improve. Therefore, the probability of looped subsets remaining after a long time is higher relative to nonlooped subsets.

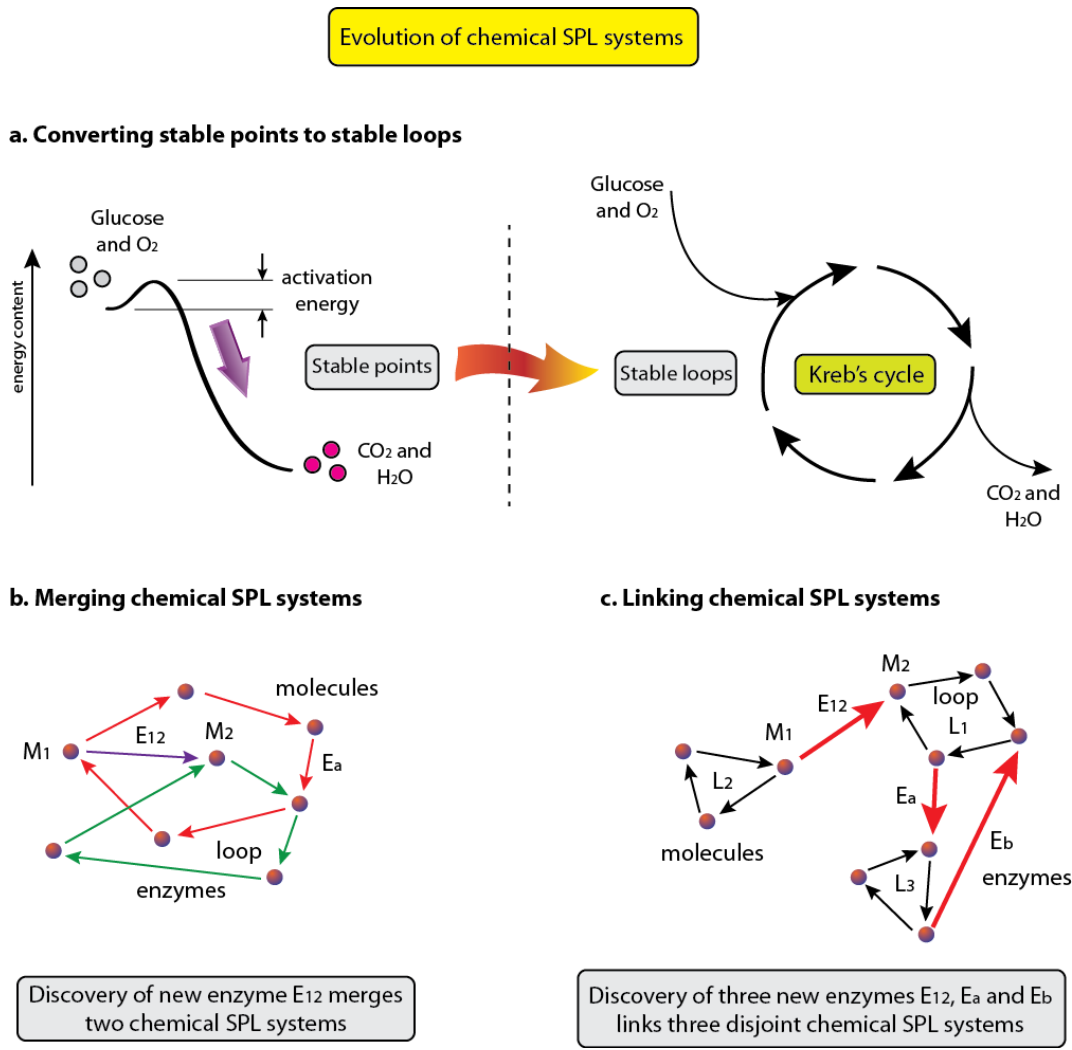

**Figure S7 | Mechanisms to create an infinite family of chemical SPL systems. a,** A chemical reaction with a stable point can be converted into a new SPL system with a chemical loop with the discovery of several new chemicals and enzymes like with Kreb's cycle. **b,** Two or more chemical SPL systems can be merged to create a new SPL system. **c,** Two or more chemical SPL systems can be linked to create a new SPL system. Using each of these processes, we create an infinite family of chemical SPL systems. Each of these processes requires the discovery of new enzymatic pathways.

If we view the static part of the chemical SPL network as a directed graph (see metabolic networks – (*11, 12*)), we can partition it into a collection of strongly connected components (*17*). Each strongly connected component has at least one loop. Such a looped network is said to be generated by a subset *S* of chemicals if and only if the entire network can be recreated using *S* and the collection of enzymes. With the above partition, the subset *S* needs to have only one chemical for each strongly connected component. Therefore, replicating the entire chemical SPL network is easy as there are several choices for subset *S* of chemicals (for example, three strongly connected components with four chemicals in each has 64 choices). A simple random split of the chemical SPL system into two parts will regenerate the entire SPL network in each part.

Even though no individual type of chemicals remains fixed or has long half-life, the entire SPL collection, as a whole, self-sustains for a long time. This is a *self-replicating property* for the entire collection. This is different from self-replication of DNA molecules as we involve the entire looped collection, not individual DNA sequences.

One special case is the problem of homochirality in which the chemicals that remain after a long time are D-form sugars and L-form amino acids. I will show how to achieve an asymmetric distribution starting from a perfectly symmetric (50-50) distribution of L- and D-forms using the chemical looped structure in a subsequent paper.

## Pumps – active material-energy looped systems

We have seen the 'existence' of a family of chemical SPL systems via a suitable mapping from physical SPL systems. However, how do we 'create' this family using natural mechanisms? I will now propose one approach using special chemical pumps, the second of the new minimal structures proposed in this paper. These mechanisms should avoid enzymes, DNA and replicases as they are highly ordered molecules, which were not yet shown to be created from a disordered collection.

Fig. S8 shows the design of a simple material-energy looped system (**Design A**). It has an elastic outer membrane, similar to the one created naturally under prebiotic conditions (*2, 3*), with no internal structure. It has one opening for inputs (as a dent) and one opening for exhaust (as a bump). The geometric shape looks approximately like a heart to facilitate suction and exhaust (discussed shortly). Inside this system, an exothermic reaction (say, a single reversible reaction) occurs. The energy released causes the membrane to expand. When it expands, more inputs are sucked in through the intake opening (Fig. S8). The elastic property of the membrane causes it to contract assuming that the system does not burst from expansion. During the contraction cycle, the outputs are pushed out through the exhaust. The exothermic reaction now proceeds once again with the new inputs. There are two loops: the physical compression-expansion cycle (at the intake and at the exhaust) and a suction-chemical reaction-exhaust outer loop. The name 'material-energy loop' or, simply, 'chemical pumps' refers to the fact that materials drive the energy creation and the energy drives the material intake, as a loop. This design clearly satisfies Definition 1 and is, therefore, an SPL system.

The pump is a simple generalization of a closed compartment with additional compression-expansion dynamics and input-output channels. The system exhibits a tendency to continue its dynamics for a long time. When we choke the intake or the exhaust, sufficient backpressure will be generated within the system from the expansion. This will try to dislodge the obstruction (like kickback in power tools). This self-sustaining ability does not apply for other damages like a tear in the membrane, as there are no loops to detect and oppose them.

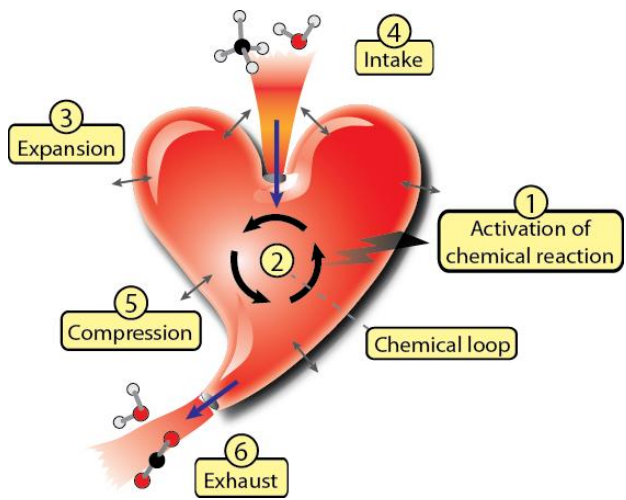

**Figure S8 | Active material-energy looped system (Design A).** When the system operates, it sucks inputs actively through the intake, expands the outer membrane, runs a chemical reaction, pushes the outputs through the exhaust and shrinks in size. The cycle repeats for a while once the system is jump-started. There are two prominent loops in this system: (a) 3-5-3 physical loop and (b) 2-3-4-5-6-2 material-energy loop.

Let me suggest a few natural variations to design A. **Design B** (Fig. S9(a)) actively fetches additional reactive chemicals like acids, bases or even ATP, if present in the environment. These secondary spontaneous chemical reactions occur in conjunction, releasing energy for the primary, possibly slower, chemical reaction. The excess energy will also make the compression-expansion cycle continue for a while. **Design C** has multiple intake and exhaust openings created through natural irregularities. In **Design D**, the intake and the exhaust openings are approximately molded to allow only a specific set of molecules (Fig. S9(b)), similar to the groves formed from erosion through repetitive operation of active water flow. **Design E** has a twist-untwist motion in addition to inward-outward motion in each compression-expansion cycle. This is common when we have nonuniform mass distribution and irregularities. As the system expands, the left-half turns counter clockwise while the right-half turns clockwise and vice versa analogous to holding a balloon with both hands, twisting it as we compress and untwisting it as we expand. Additional variations are combinations of the above designs and other active mechanisms that use just the above abstract features.

In the above designs, even though I have assumed the reactions to be exothermic that generate more energy than what the chemical and physical parts of the system consumes, there are two ways to relax these assumptions. The first way is to use the excess energy available in the environment like within hydrothermal vents or in a hot primitive earth. The second way is to use design B in which a secondary exothermic reactive reaction is coupled with an endothermic primary reaction.

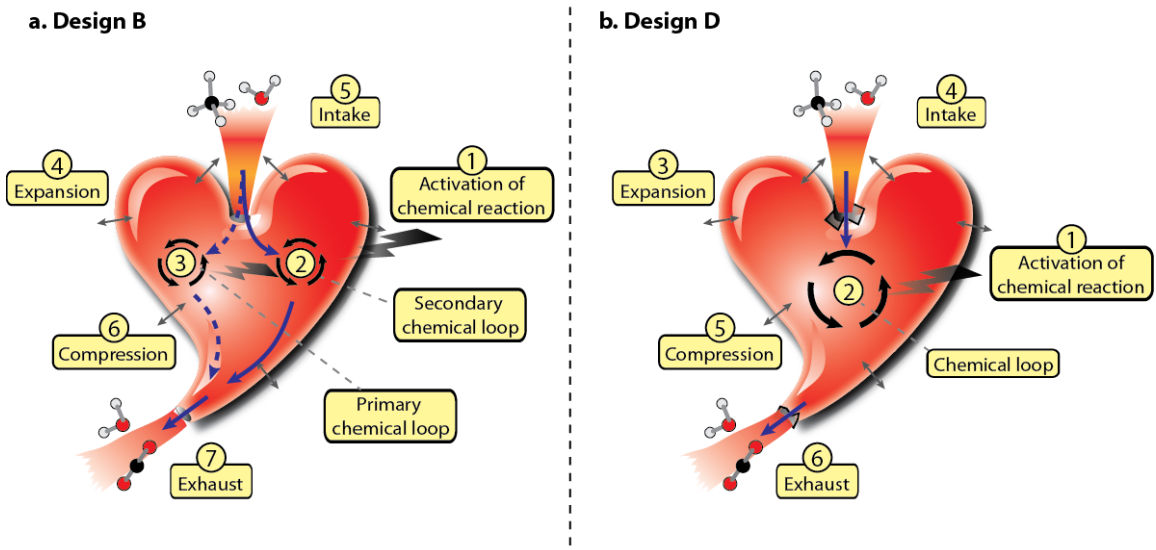

**Figure S9 | Variations to active material-energy looped system. a,** Design B is a variation in which a secondary chemical loop powers the primary chemical loop to execute its dynamics. **b,** Design D is a variation in which the intake and exhaust openings are altered to allow only special shaped and sized molecules.

What are the chances of creating the structure and running the dynamics naturally? If we include help from conscious humans, we already have efficient designs like internal combustion engines, pumps with similar intake and exhaust openings and nanotube nanomotors (*18, 19*). Even without such help, it has been shown experimentally that vesicles (i.e., closed compartments) do form under primitive earth conditions using fatty acids, organized as micelles (*2, 3*), instead of relying on phospholipids. It was also shown in Szostak Lab that clay montmorillonite catalyzes the vesicle formation and that the vesicles are stable for days to months. In addition, they have shown that when fatty acid micelles are added, the vesicles even grow and split using a process called flipping (a primitive reproduction mechanism for just the membrane).

The designs A-E proposed here are simple and do not require anything else other than the outer membrane with, say, two irregularities (openings for intake and exhaust). Therefore, the current experimental results from Szostak Lab (*2, 3*) already show the 'structural' feasibility of these designs under primitive conditions. The reproduction mechanism of flipping is sufficient to recreate these designs repeatedly for a long time.

However, the above designs are useful only when they execute their dynamics. Recently, catalytic nanomotors (*19*) that have self-propulsion and functionally similar properties to pumps were built by inserting carbon nanotubes into gold and platinum nanowires synthetically to achieve high speeds (50-60 μm/s). They even combined hydrogen peroxide fuel with hydrazine to achieve higher speeds (94-200 μm/s) comparable to natural biomolecular motors. However, creating functional pumps under natural primitive conditions is not yet done.

If the material-energy looped system were to continue the dynamics for a while, the following conditions must be satisfied: (a) the looped process should be initiated by jump-starting the system, (b) the energy generated should be more than the energy consumed for the compression-expansion cycle and other losses, (c) the inputs should not run out, (d) the intake and exhaust should not be severely blocked and (e) there should be no damages to the system. These conditions lower the probability for the dynamics, though they continue to occur repeatedly.

Most existing molecular-sized pumps in living beings (not organ-sized pumps like heart) like sodium-potassium pumps, calcium pumps, proton pumps, large class of ABC pumps and others, use specialized proteins. They are efficient, as are synthetic nanomotors (*19*). In comparison, the pumps suggested here only need to be feasible with a small enough efficiency of operation. This is reasonable owing to its simplicity, but needs experimental verification. The likely locations for these designs to occur naturally would align with geological evidence like within hydrothermal vents and hot springs.

## Primitive catalysts

How can we use the above pumps, which are simple SPL systems, to create more complex chemical looped systems? The answer lies in recognizing that the pump designs A-E behave functionally as catalysts, similar to enzymes in a cell. This would allow us to replace the entire family of chemical SPL designs that relied on enzymes (Figs. S5-S7) with ones that use the active pumps.

To see the catalytic property, note first that any chemical reaction occurring passively in the environment through random chance, self-assembly or even within a sealed flask in a lab can now occur within the enclosed pumps proposed here. Secondly, the rate of reaction is much faster using these designs (due to active suction mechanism, confinement within the membrane and accumulation of inputs and energy through repeated cycles, better temperatures, pressures and other conditions) compared to the alternative of occurring openly in the environment. Thirdly, the designs themselves do not undergo any change while running the internal chemical reaction.

These properties together define a catalyst. These are universal catalysts as they are not specialized to a given chemical reaction (unlike enzymes). The same pump can run one chemical reaction now and a different reaction a few minutes later. If we use design D instead, in which the shape of the intake and exhaust openings are morphed to allow and even assist a specific set of molecules to enter or leave during the compression-expansion cycle, the design becomes analogous to the induced-fit model of an enzyme (*20*). The pumps continuously and autonomously draw inputs and push outputs for a long time as an active looped process. This operation, though simple and is theoretically feasible, would need to be verified experimentally.

Each chemical SPL system from the infinite family identified in the previous section were composed of several uphill and downhill reactions. The pumps, now acting as primitive catalysts, are capable of running each of these reactions, albeit slowly and inefficiently. More generally, the same infinite family of chemical SPL systems in the previous section can now be modified to use pumps instead of using enzymes (for the inclined plane in the mapping), at least for reactions involving simple non-polymer molecules.

If we have millions of these pumps in a sealed room, the outputs from the pumps will be taken as inputs to generate new chemicals iteratively. This is another stable looped process that continues to occur within the sealed room. As a result, we begin to see new collection of chemicals beyond the initial random collection we started with. The new chemicals and the pumps acting as catalysts will start to form chemical SPL networks. This system exhibits a self-replicating property. This is the beginning of generation of order from disorder in which random chemical collections begin to lock themselves to form looped networks.

The active pumps are capable of creating random polymer sequences like random proteins using amino acid monomers, random RNA and DNA sequences using nucleotide monomers and random polymers based on glucose (like cellulose and starch), other monosaccharides and lipids. Creating correlated DNA and RNA sequences from random sequences will be considered as future work (using pumps, helical geometry and natural affinity between adenine-thymine/uracil and guanine-cytosine) along with experimental justification.

## Conclusions

The new theoretical framework introduced in this paper identifies two new minimal universal structures and uses them to *construct* an infinite family of dynamical systems called stable parallel looped systems (SPL) with any specified complexity. No other class of dynamical systems guarantees stability for a long time when a system has millions of interconnected moving parts. The new SPL framework provides the hidden structures within dynamical systems to create increasingly complex ordered systems from disorder. In contrast, all existing theories rely on passive and random chance-based mechanisms for the evolution of order. These are unconvincing for systems as complex as living beings, especially since they do not address the difficulties associated with the second law of thermodynamics. Active SPL systems are the patterns and structures necessary to drive the creation of order from disorder.

As an example, just as *individual* molecules lock themselves into *stable configurations* under any given environmental conditions and at any given time (like $H_2O$ is more stable than $H_2O_2$ under current conditions), a *collection of different* types of molecules lock themselves into *stable loops* at all times (Claim 1). This collection as a whole becomes a universal structure that guarantees the creation of order within a disorderly collection and arrangement of molecules.

How can we keep increasing the stable looped order favorably over time in spite of the disorder? Two necessary processes are (a) by constantly creating new and diverse set of chemical collections – the more the diversity of organic chemicals, the more interconnected is the resulting looped network that the molecules get locked into and (b) by continuously regenerating the entire chemical collection discovered until then – a self-replicating property applied to the collection. Nature cannot rely on chemicals created only once, by chance. For the former, we need simple but *active* pumps (designs A-E, see Figs. S8 and S9). For the latter, the looped dynamical property first ensures repeatability of creation of monomers and simple molecules. The extension of these theoretical mechanisms to polymers will be considered in the future. The designs proposed here are simple and are theoretically feasible. Furthermore, experimental justification of *physical* SPL systems is already well-known (cf. roller coaster designs). However, experimental justification with *chemical* SPL systems using pumps is not yet done and is an important area of future work.

# Acknowledgements

I thank my wife Sridevi Ravuri for over ten years of discussions, comments and for the emotional support.